\documentclass[11pt]{article}
\usepackage{pgfplots}

\usepackage[preprint]{acl}

\usepackage{times}
\usepackage{latexsym}
\usepackage{amssymb}
\usepackage{amsmath}
\usepackage{tikz}
\usepackage{pgf-pie}
\usepackage{multirow}
\usepackage{bbm}
\pgfplotsset{compat=1.18}
\usepackage[T1]{fontenc}

\usepackage[utf8]{inputenc}

\usepackage{microtype}

\usepackage{inconsolata}

\usepackage{graphicx}

\newcommand\blfootnote[1]{%
  \begingroup
  \renewcommand\thefootnote{}\footnote{#1}%
  \addtocounter{footnote}{-1}%
  \endgroup
}

\title{WINDQuant: Weight-Informed Neural Decision-Making for Global Mixed-Precision LLM Quantization}

\author{
\textbf{Phong Nam Huu Nguyen}$^{1,2}$ \quad
\textbf{Khoi M. Le}$^{4}$ \quad
\textbf{Cong-Duy T Nguyen}$^{1}$ \\
\textbf{Anh Tuan Luu}$^{1, 3}$ \quad
\textbf{Thong Thanh Nguyen}$^{\dagger 4}$ \quad
\textbf{Tho Quan}$^{2}$ \\
\\
$^{1}$CAIR, VinUniversity, Vietnam \\
$^{2}$Ho Chi Minh City University of Technology (HCMUT), VNU-HCM, Ho Chi Minh City, Vietnam \\
$^{3}$CCDS, Nanyang Technological University, Singapore \\
$^{4}$School of Computing, National University of Singapore, Singapore \\
\texttt{phong.nhn@vinuni.edu.vn, thong.nguyen@u.nus.edu}
}

\begin{document}
\maketitle
\begin{abstract}
Quantization is an effective approach to reduce the memory footprint and inference cost of large language models (LLMs), yet maintaining performance in the ultra-low-bit regime remains challenging. Existing post-training methods often suffer from severe accuracy degradation, while quantization-aware training requires costly retraining and additional resources. Moreover, most mixed-precision strategies rely on coarse-grained or heuristic sensitivity analysis that overlooks fine-grained variations within weight matrices. We propose WINDQuant, a reinforcement-learning-based allocation controller for ultra-low-bit LLM quantization. Rather than introducing another low-level quantization operator, WINDQuant learns how to assign bit-widths and quantization treatments to fine-grained column chunks under a global storage budget. By operating at the column-chunk level, WINDQuant enables flexible and fine-grained precision assignment within layers under a global target bit-width. The implementation combines PPO with activation-aware calibration, lightweight per-unit quantizer fitting, and explicit effective-bit accounting of the learned mixed-precision plan. Experiments on LLaMA models demonstrate that WINDQuant achieves competitive performance in ultra-low-bit settings while reducing optimization overhead relative to retraining-based approaches, highlighting reinforcement learning as a practical controller for adaptive mixed-precision quantization.
\blfootnote{$^{\dagger}$Corresponding author.}
\end{abstract}

\section{Introduction}
Large Language Models (LLMs) have achieved remarkable performance across reasoning, language understanding, and generation tasks \cite{brown2020language,chowdhery2023palm}. However, this progress comes at the cost of massive computational and memory requirements. Modern frontier models contain hundreds of billions to trillions of parameters, requiring terabytes of VRAM for full-precision deployment \cite{fedus2022switch}. Such resource demands make direct deployment on edge devices or resource-constrained environments infeasible. Quantization has emerged as one of the most promising solutions to reduce inference cost by representing model weights and activations with lower-precision data types \cite{dettmers2022gpt3}.

Despite its success, low-bit quantization remains highly challenging, especially in the sub-3-bit regime \cite{banner2019post}. Post-Training Quantization (PTQ) methods are efficient and fast but tend to suffer severe degradation at ultra-low precision \cite{nagel2019data}. Quantization-Aware Training (QAT) methods improve robustness but require expensive retraining and often rely on knowledge distillation, significantly increasing computational overhead \cite{jacob2018quantization}. Moreover, most existing approaches apply uniform or coarse-grained quantization strategies, overlooking the fact that not all weights contribute equally to model performance \cite{dong2019hawq}.

Recent empirical findings indicate that a small fraction of salient weights disproportionately affects model accuracy \cite{lin2024awq}. Furthermore, sensitivity exhibits structured locality across layers and channels, suggesting that different substructures within a layer respond differently to quantization \cite{dong2019hawq}. This observation motivates a more fine-grained and adaptive allocation strategy. Instead of treating each tensor, group, or unit as an independent local quantization problem, we view ultra-low-bit mixed-precision quantization as a global budgeted allocation problem, where assigning more precision to one unit necessarily reduces the budget available to others.

In this work, we introduce WINDQuant (Weight-Informed Decision-making for Reinforcement Learning in Quantization), a reinforcement learning framework for mixed-precision quantization. We define a unit as a column-level chunk within a layer and formulate quantization as a finite-horizon sequential decision problem with a global target bit budget. At each step, the agent observes statistical, structural, and activation-aware features of a unit and selects one of seven actions: skip or quantize at 8, 4, 3, 2, 1.58, or 1 bit. The training objective combines PPO \cite{schulman2017proximal} with budget-aware action masking, dense step-level shaping, and an episode-end quality penalty computed from a fast perplexity proxy. This distinction is central to our formulation. Conventional PTQ and QAT methods primarily specify how a given tensor, group, or block should be quantized once the target precision is fixed. WINDQuant instead addresses the complementary allocation problem: given a pretrained model decomposed into thousands of sub-layer units, the policy decides which units should be kept at higher precision, aggressively quantized, or assigned ternary/binary operators under a global bit budget.

Unlike prior RL-based approaches that operate at layer granularity \cite{wang2019haq}, WINDQuant performs fine-grained unit-level allocation, enabling more expressive and adaptive mixed-precision policies. Our framework integrates activation-aware protection mechanisms and supports multiple quantization operators spanning 1-bit to 8-bit regimes. Experiments on LLaMA models demonstrate that WINDQuant achieves competitive performance in ultra-low-bit settings while significantly reducing computational and memory overhead compared to training-based approaches.

Overall, the contributions of our research can be articulated as follows:
\begin{itemize}
    \item We propose \textbf{WINDQuant}, a reinforcement learning--based framework that reformulates fine-grained mixed-precision quantization for large language models as a sequential decision-making problem, enabling adaptive bit-width allocation beyond static heuristic sensitivity ranking.
    
    \item We demonstrate the effectiveness of the proposed approach across a range of modern large language models, including LLaMA3 (1B, 3B, 8B, 70B), showing its generality and applicability to contemporary architectures.
    
    \item We show that WINDQuant reaches competitive ultra-low-bit performance on LLaMA models up to 70B parameters while avoiding full model retraining, requiring only lightweight policy optimization and local per-unit quantizer fitting.
\end{itemize}

\section{Related Works}

\subsection{Quantization of Large Language Models}
Quantization methods for LLMs are commonly divided into post-training quantization (PTQ) and quantization-aware training (QAT) \cite{lang2024comprehensive, li2024contemporary}. PTQ approaches compress pretrained models without additional training by applying calibration and reconstruction techniques \cite{lin2024awq, xiao2023smoothquant}. Their primary strength lies in efficiency, but performance degrades significantly in ultra-low-bit settings, where fixed or heuristic precision assignments cannot adequately compensate for accumulated quantization error. QAT methods address this by incorporating quantization effects during training \cite{liu2024llm}, yielding stronger accuracy retention at the cost of substantial retraining overhead. Despite their complementary strengths, both paradigms typically treat precision allocation as a static optimization problem, motivating adaptive approaches that learn allocation policies automatically.

\subsection{Reinforcement Learning for Neural Network Compression}
Reinforcement learning has emerged as a promising framework for automated model compression, framing design decisions as sequential optimization problems. HAQ \cite{wang2019haq} pioneered the use of deep deterministic policy gradient (DDPG) to determine layer-wise bit-widths for hardware-aware mixed-precision quantization without manual heuristics. ReLeQ \cite{elthakeb2020releq} applied Q-learning to learn heterogeneous bit-width assignments, demonstrating that RL-based policies can approach near-lossless accuracy across different architectures. Beyond quantization, AMC \cite{he2018amc} uses DDPG to automate channel pruning ratios, and N2N learning \cite{ashok2018n2n} applies policy gradient methods to jointly optimize architecture and compression decisions. These works share a common insight: compression decisions involve complex, interdependent trade-offs that are naturally suited to sequential decision-making formulations.

However, existing RL-based quantization methods face several limitations when applied to large language models. First, operating at layer granularity leads to coarse allocation that ignores within-layer sensitivity variations. Second, reward evaluation typically requires repeated full-model inference, creating computational bottlenecks at the billion-parameter scale. Third, the action spaces and state representations used in prior work were designed for smaller vision models and do not account for the structural properties of transformer-based LLMs, such as the distinct sensitivity profiles of attention and MLP components.

These limitations motivate our approach: a fine-grained, sub-layer RL formulation with efficient reward evaluation, specifically designed for large-scale transformer quantization. Unlike HAQ and ReLeQ, which assign one bit-width per layer, WINDQuant operates at the column-chunk level, enabling more precise adaptation to within-layer sensitivity.

\subsection{Mixed-Precision and Sensitivity-Aware Quantization}
Empirical findings show that only a small fraction of salient weights dominate model accuracy, with sensitivity structured along channels rather than uniformly distributed \cite{lin2024awq, dong2019hawq}. Recent elastic quantization operators--such as Learned Step-Size Quantization (LSQ) and Stretched Elastic Quantization (SEQ)--adaptively learn per-group quantization boundaries, outperforming fixed-scale methods at ultra-low precision \cite{LiuZechun2025PISL}. Most existing mixed-precision approaches apply coarse layer-level or mask-based heuristics that ignore fine-grained within-layer variation \cite{li2024contemporary}. These operators serve as quantization primitives within our RL framework, enabling the agent to leverage optimized quantizers at each precision level.

\section{Methodology}

\subsection{Problem Definition}
Let $\mathcal{M}$ denote a pretrained model with $N$ quantization units. Each weight matrix is decomposed into column-level chunks, referred to as \emph{units}, motivated by empirical observations that weight importance exhibits channel-localized structure \cite{lin2024awq}. Each unit $i$ has $n_i$ weights and is assigned an action $a_i$. Because the implementation allows skip decisions and INT8-protected weights, the realized storage cost is an effective bit count $B_i(a_i)$ rather than just a nominal action bit-width. We seek a policy $\pi$ that satisfies a global storage budget:

\begin{equation}
  \label{eq:budget}
        \sum^{N}_{i=1}{B_i(a_i)} \leq B_{\mathrm{target}}
\end{equation}

while preserving model quality under a soft constraint:

\begin{equation}
  \label{eq:quality}
    \mathcal{L}(\mathcal{M}_Q; \mathcal{D}) - \mathcal{L}(\mathcal{M}; \mathcal{D}) \leq \varepsilon
\end{equation}

where $\mathcal{M}_Q$ is the quantized model and $\mathcal{D}$ is the calibration corpus used during RL reward evaluation. Downstream reasoning benchmarks are used only after training. Rather than assigning bit-widths independently per layer using local heuristics, we formulate this as a sequential decision process where the policy considers global compression state and previously made decisions, enabling adaptive allocation under a global budget.

\subsection{MDP Formulation}
We cast the bit-width assignment problem as a finite-horizon MDP $(\mathcal{S}, \mathcal{A}, P, R, \gamma)$.

\subsubsection{State Space}
The state at step $t$ encodes both local weight statistics and global compression context. Table~\ref{table1} details the 15-dimensional state representation. Following prior RL-based compression methods \cite{elthakeb2020releq, wang2019haq}, we include distributional features of the current weight unit (mean, variance, sparsity, outlier fraction) to estimate quantization sensitivity. The state also includes episode-level signals such as progress, remaining bit budget, and urgency, allowing the agent to adapt its strategy as the compression budget becomes constrained. Calibration statistics derived from activation-aware analysis provide information about activation scales, which correlate with weight importance. Structural indicators help the policy distinguish between attention and MLP components, which exhibit different sensitivity to quantization.

\begin{table}
  \centering
  \begin{tabular}{c l l}
    \hline
    \textbf{Dim.} & \textbf{Feature} & \textbf{Formula} \\
    \hline
    0  & mean & $\bar{W} $ \\
    1  & std & $\sigma_W $ \\
    2  & abs\_mean & $\frac{1}{n}\sum_i |W_i|$ \\
    3  & sparsity & $\frac{|\{i : |W_i| < 10^{-6}\}|}{n}$ \\
    4  & outlier\_frac & $\frac{|\{i : |W_i| > 0.5 \cdot P_{99}\}|}{n}$ \\
    5  & progress & $\frac{t}{N}$ \\
    6  & bits\_used\_ratio & $\frac{B_{used}}{B_{{FP32}}}$ \\
    7  & budget\_remaining & $\frac{B_{{target}} - B_{{used}}}{B_{{target}}}$ \\
    8  & urgency & $(1-\frac{B_r}{b_t n_r})_+$ \\
    9-11 & calib\_stats & $(\bar{s}, \sigma_s, \max(s))$ \\
    12 & is\_attention & ${1}[{``attn"} \in {name}]$ \\
    13 & is\_mlp & ${1}[{``mlp"} \in {name}]$ \\
    14 & target\_norm & $b_{{target}} / 8$ \\
    \hline
  \end{tabular}
  \caption{State feature definitions used in WINDQuant.}
  \label{table1}
\end{table}

In the implementation, budget-related state features are normalized by an FP32 reference total ($32\sum_i n_i$) for numerical stability, even though realized storage may include FP16 skip actions and INT8-protected weights.

\subsubsection{Action Space}
The action space defines the set of discrete precision levels that the agent can assign to each quantization unit:

\begin{equation}
    \label{eq:action-space}
    \mathcal{A} = \{16;8;4;3;2;1.58;1\}
\end{equation}

The actions correspond to supported bit-widths of the quantization engine, ranging from full-precision (16-bit),  conventional low-precision formats (8- and 4-bit) to ultra-low-bit representations (3-, 2-, and 1-bit). We additionally include the 1.58-bit format, which corresponds to ternary quantization schemes. By allowing the agent to choose among these discrete precision levels, the policy can adaptively allocate higher precision to sensitive units while aggressively compressing less critical components.

In practice, the categorical policy is combined with budget-aware action masking, and the selected action is further clamped by layer-specific minimum precisions for early layers, attention projections, and MLP/gate blocks.

\paragraph{Episode complexity.}
For LLaMA-3.1-8B with column chunk size $G{=}256$, each episode processes $N{\approx}3{,}584$ units (32 layers $\times$ 7 linear projections per layer $\times$ 16 chunks per projection). Reward evaluation uses 64 Open-Platypus samples truncated to sequence length 256 and serves as a fast training proxy rather than a full downstream benchmark pass. In our saved 8B runs, a 100-episode training job completes in tens of hours on a single H200 GPU.

\subsubsection{Transition and Discount}
The transition is deterministic: given the current state and action, the environment applies the selected quantization to the current unit and advances to the next. The discount factor is set to $\gamma = 0.99$.

\subsection{Reward Design}
\label{sec:reward}

The mixed-precision allocation problem balances compression against proxy quality under a target average bit-width. The total episode return combines local per-step feedback with an episode-end proxy evaluation:

\begin{equation}
    \label{eq:reward-total}
    G = \sum_{t=1}^{T} \gamma^{T-t} r_{\mathrm{step}}(t) + R_{\mathrm{final}}
\end{equation}

where $r_{\mathrm{step}}(t)$ nudges the policy toward budget-feasible local decisions and $R_{\mathrm{final}}$ scores the realized mixed-precision plan. This hybrid structure mitigates the credit assignment problem without replacing the terminal objective.

\paragraph{Episode-end reward.} The final reward balances compression efficiency against model quality through a weighted combination of objectives:

\begin{equation}
\label{eq:reward}
\mathcal{R} = \alpha G_c - P_{\mathrm{qual}} - P_{\mathrm{budget}} - P_{\mathrm{ppl}} - P_{\mathrm{skip}} + B_{\mathrm{quality}}
\end{equation}

where $G_c$ measures normalized bit reduction relative to an FP32 reference and is capped to account for INT8 protection overhead; $P_{\mathrm{qual}}$ penalizes relative loss increase beyond a tolerance threshold $\varepsilon{=}0.3$, modulated by a Lagrange multiplier $\lambda$; $P_{\mathrm{budget}}$ penalizes violation of the target bit-width with a super-linear penalty; $P_{\mathrm{ppl}}$ imposes a piecewise penalty on the perplexity ratio $\rho = \mathrm{PPL}_Q / \mathrm{PPL}_0$ using soft and hard thresholds of 3 and 20; $P_{\mathrm{skip}}$ discourages excessive skipping beyond 10\% of units; and $B_{\mathrm{quality}}$ provides an additional bonus when the perplexity ratio remains below 2.

\paragraph{Per-step reward.} To provide dense temporal feedback, step-level rewards use lightweight shaping signals: skip actions incur a small penalty, allocations at or below the current target receive a positive signal proportional to saved bits, and over-target actions receive a mild negative reward. The dominant learning signal still comes from the episode-end reward.

\paragraph{Constraint enforcement.} We handle the quality constraint (Eq.~\ref{eq:quality}) via Lagrangian relaxation: a dual variable $\lambda$ is updated by projected gradient ascent after each episode, increasing the quality penalty weight when degradation is excessive and relaxing it otherwise. The bit budget is not enforced by the dual variable; instead, the implementation combines budget-aware action masks with an episode-end budget penalty. Additionally, we adopt a curriculum learning strategy where the target bit-width is progressively reduced (e.g., $3.0 \to 2.5 \to 2.0 \to 2.0$). This staged optimization stabilizes exploration and allows the policy to learn meaningful quantization patterns before aggressive compression is applied.

\subsection{Policy Optimization}
\label{ss:ppo}

We optimize the precision-allocation policy using Proximal Policy Optimization (PPO) \cite{schulman2017proximal}, a stable first-order policy gradient method. The policy $\pi_\theta(a \mid s)$ is parameterized by a neural network that maps state representations to a categorical distribution over bit-width actions. In the implementation, the actor-critic network is a two-layer MLP with LayerNorm and ReLU activations, followed by separate actor and critic heads over the seven-way action space.

PPO's clipped surrogate objective prevents excessively large policy updates by constraining the probability ratio between the updated and previous policies, approximating a trust-region constraint without the computational cost of second-order methods. Advantages are estimated using Generalized Advantage Estimation (GAE) \cite{schulman2015high} with $\gamma = 0.99$ and $\lambda_{\mathrm{GAE}} = 0.95$, which provides smooth credit assignment across sequential bit-width decisions--critical since early allocation decisions affect later budget feasibility.

The full optimization objective combines the clipped surrogate loss, value function regression ($c_1 = 0.5$), and entropy regularization ($c_2 = 0.05$). In the saved v4 runs, PPO performs 3 optimization epochs per roll-out. The entropy term discourages premature convergence to deterministic bit-width assignments, which is important because early decisions influence remaining budget flexibility and final model quality.


\subsection{Quantization Engine and Bit Accounting}
\label{subsec:engine}

After the policy selects an action for a column chunk, the quantization engine applies the corresponding low-bit operator: elastic binarization for 1-bit weights, SEQ-style operators for 1.58/2-bit weights, and LSQ-style sparse quantization for 3/4/8-bit weights. Each operator performs lightweight local scale fitting on the current unit without backpropagating through the full model. If the reconstruction error of a selected operator is excessive, the implementation falls back to a higher-precision operator. Additional implementation details are provided in Appendix~\ref{appdx:engine}.

We also use activation-aware salient weight protection \cite{lin2024awq}. For each unit, a small fraction of salient weights is identified using the product of weight magnitude and activation scale and stored in INT8; the remaining weights use the bit-width selected by the policy. Therefore, the realized storage cost differs from the nominal action bit-width. For a unit $i$ with $n_i^{(p)}$ protected weights and $n_i^{(q)}=n_i-n_i^{(p)}$ quantized weights, the effective bit cost is
\begin{equation}
\label{eq:effective-bits}
B_i =
\begin{cases}
16 \cdot n_i, & \text{if action = 16}, \\
b_i \cdot n_i^{(q)} + 8 \cdot n_i^{(p)}, & \text{otherwise}.
\end{cases}
\end{equation}
All reported average bit-widths use $\bar{b}=\sum_i B_i/\sum_i n_i$, so the reported compression level includes the overhead of INT8-protected weights.

\begin{figure*}
    \centering
    \includegraphics[width=1.0\linewidth]{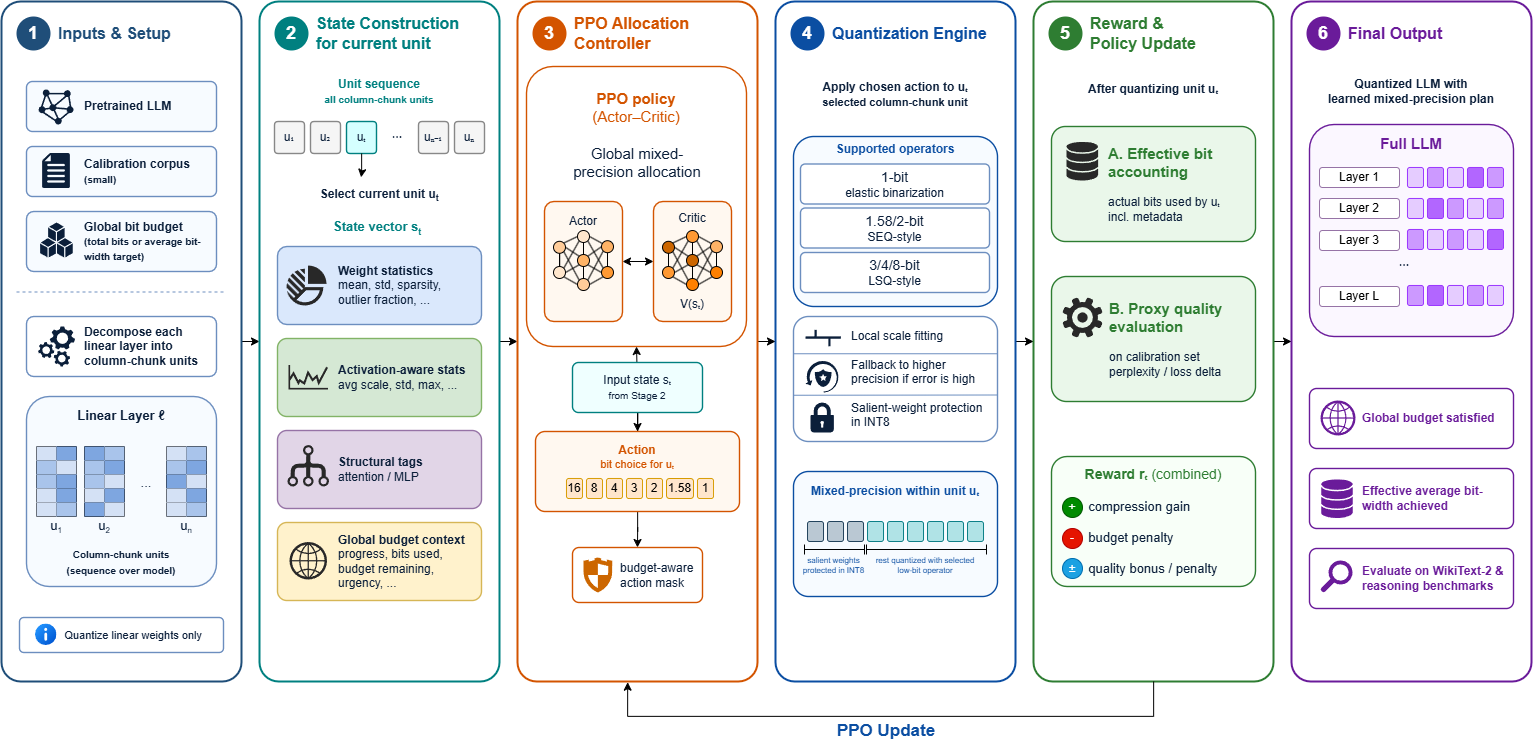}
    \caption{Visualization of the WINDQuant pipeline within a single quantization step.}
    \label{fig:pipeline}
\end{figure*}

In summary, WINDQuant should be understood as an end-to-end allocation-and-quantization pipeline, as illustrated in Figure~\ref{fig:pipeline}.

\section{Experiments}

In this section, we evaluate WINDQuant across a range of LLaMA-3 family models to assess its ability to balance reasoning performance, language modeling quality, and computational efficiency under ultra-low-bit mixed-precision settings. Our method is compared against PTQ approaches--AWQ \cite{lin2024awq}, GPTQ \cite{frantar2022gptq}, 
OmniQ \cite{shao2023omniquant}, SpinQuant \cite{liu2024spinquant}, 
TesseraQ \cite{LiYuhang2024TULL}, GPTVQ \cite{vanBaalenMart2024GTBo}, SignRound \cite{ChengWenhua2024OWRv}
--QAT methods--LLM-QAT \cite{liu2024llm}, EfficientQAT \cite{chen2025efficientqat},
--and vector quantization (AQLM \cite{egiazarian2024extreme}).

\subsection{Experiment Settings}
We evaluate WINDQuant on LLaMA-3.2-1B, LLaMA-3.2-3B, LLaMA-3.1-8B, and LLaMA-3-70B. All models are quantized to ultra-low bit-width, approximately 2-bit mixed precision. 

For the main tables, we evaluate language modeling quality and reasoning performance using the WikiText-2 (Wiki2) test set \cite{merity2016pointer} and zero-shot commonsense reasoning benchmarks, including ARC-E, ARC-c \cite{clark2018think}, BoolQ \cite{clark2019boolq}, PIQA \cite{bisk2020piqa}, HellaSwag \cite{zellers2019hellaswag}, and Winogrande \cite{sakaguchi2021winogrande}. All experiments are conducted on a single NVIDIA H200 GPU (140GB VRAM). We report macro-average accuracy across tasks, giving each benchmark equal weight. For baseline evaluation, we reproduce all feasible methods using official implementations or widely adopted codebases, following hyperparameters recommended by the corresponding papers and repositories. All resulting models are evaluated with the same downstream benchmark pipeline. We do not apply additional method-specific tuning beyond official recommendations, since our goal is to compare methods under their intended usage rather than engineer custom rescue configurations for individual baselines. Full baseline configurations are provided in Appendix~\ref{sec:baseline_configs}.

During quantization, all models are initialized from pretrained checkpoints. Following standard practice, we quantize linear weights while excluding embedding and output layers to preserve stability. Unless otherwise stated, WINDQuant uses column chunks of size 256, 64 Open-Platypus calibration samples truncated to sequence length 256, 10 local quantizer-optimization steps per unit, and PPO with 3 epochs per roll-out, learning rate $5 \times 10^{-4}$, clipping parameter 0.2, and entropy coefficient 0.05. 

Training budgets are model-dependent in the saved runs used for this paper. The main 1B, 3B and 8B runs use 100 episodes with curriculum $\{3.0, 2.5, 2.0, 2.0\}$, while 70B runs use 24 episodes for runtime reason. Activation-aware protection stores salient weights in INT8; depending on the run, fixed salient rates are set at 3\%. During training, the reward is computed on the Open-Platypus proxy corpus; downstream benchmarks are used only for post-training evaluation.

\subsection{Results \& Discussion}

\begin{table*}[t]
\centering
\small
\setlength{\tabcolsep}{3.5pt}
\begin{tabular}{c|l|c|cccccc|c}
\hline
\textbf{Model} & \textbf{Method} & \textbf{\#Bits} 
& \textbf{ARC-c} $\uparrow$ 
& \textbf{BoolQ} $\uparrow$ 
& \textbf{PIQA} $\uparrow$ 
& \textbf{HS} $\uparrow$ 
& \textbf{Wino} $\uparrow$ 
& \textbf{Avg} $\uparrow$ 
& \textbf{Wiki2} $\downarrow$ \\
\hline

\multirow{11}{*}{1B}
& FP            & 16   & 36.7 & 63.8 & 74.8 & 64.2 & 60.7 & 60.04 & 8.62 \\
\cline{2-10}
& WINDQuant     & 2.01 & \textbf{28.0} & 52.7 & \textbf{64.3} & \textbf{47.0} & \textbf{54.5} & \textbf{49.30} & \textbf{19.64} \\
& AWQ           & 2    & 26.0 & 50.7 & 52.2 & 26.0 & 50.7 & 41.12 & $1.7\times10^5$ \\
& GPTQ          & 2    & 25.0 & 41.2 & 50.6 & 26.9 & 51.0 & 38.94 & $1.2\times10^5$ \\
& LLM-QAT       & 2    & 25.2 & 37.8 & 52.3 & 25.6 & 49.1 & 38.00 & 6260 \\
& OmniQ         & 2    & 26.2 & 37.8 & 53.6 & 25.9 & 48.9 & 38.48 & 7945 \\
& SpinQuant     & 2    & 25.7 & 44.0 & 50.7 & 26.5 & 51.5 & 39.68 & 56.34 \\
& EfficientQAT  & 2.12 & 24.3 & \textbf{60.3} & 59.8 & 35.7 & 52.8 & 46.58 & 29.32 \\
& GPTVQ         & 2    & 23.8 & 45.9 & 55.7 & 31.2 & 50.6 & 41.44 & 50.93 \\
& SignRound     & 2    & 26.0 & 53.2 & 59.8 & 33.3 & 53.0 & 45.09 & $4.96\times10^4$ \\
& TesseraQ      & 2    & 23.3 & 40.4 & 52.8 & 30.0 & 50.4 & 39.38 & 30.44 \\
\hline

\multirow{12}{*}{3B}
& FP            & 16   & 46.3 & 74.1 & 78.2 & 74.1 & 69.5 & 67.61 & 6.93 \\
\cline{2-10}
& WINDQuant     & 1.99 & \textbf{37.8} & 66.8 & \textbf{73.9} & \textbf{65.7} & \textbf{65.7} & \textbf{61.98} & \textbf{10.40} \\
& AWQ           & 2    & 25.5 & 37.8 & 52.9 & 26.3 & 50.2 & 38.54 & $6.6\times10^4$ \\
& GPTQ          & 2    & 24.0 & 43.7 & 51.3 & 27.7 & 47.7 & 38.88 & $4.7\times10^4$ \\
& LLM-QAT       & 2    & 25.7 & 37.8 & 50.7 & 41.5 & 54.4 & 42.02 & 6471 \\
& OmniQ         & 2    & 24.2 & 37.8 & 50.7 & 26.1 & 48.9 & 37.54 & 9657 \\
& SpinQuant     & 2    & 27.0 & 46.2 & 52.4 & 26.4 & 47.9 & 39.98 & 57.40 \\
& EfficientQAT  & 2.12 & 24.8 & 43.0 & 64.1 & 37.1 & 55.0 & 44.80 & 21.49 \\
& AQLM          & 2.27 & 37.6 & \textbf{70.5} & 73.8 & 64.4 & 63.0 & 61.86 & 11.29 \\
& GPTVQ         & 2    & 20.7 & 37.9 & 52.4 & 27.9 & 49.8 & 37.74 & 17.57 \\
& SignRound     & 2    & 31.3 & 61.5 & 67.4 & 48.9 & 56.5 & 53.13 & $1.01\times10^5$ \\
& TesseraQ      & 2    & 24.6 & 54.6 & 60.1 & 37.3 & 54.1 & 46.13 & 30.22 \\
\hline

\multirow{12}{*}{8B}
& FP            & 16   & 55.03 & 83.05 & 81.12 & 79.3 & 74.58 & 74.62 & 5.56 \\
\cline{2-10}
& WINDQuant     & 1.96 & \textbf{49.9} & \textbf{77.1} & \textbf{78.4} & \textbf{75.4} & 70.3 & \textbf{70.22} & 7.90 \\
& AWQ           & 2    & 25.7 & 42.9 & 50.3 & 25.9 & 48.5 & 38.66 & $2.8\times10^5$ \\
& GPTQ          & 2    & 24.2 & 50.5 & 52.7 & 30.2 & 50.4 & 41.60 & 10092 \\
& LLM-QAT       & 2    & 21.6 & 37.8 & 50.3 & 26.1 & 49.9 & 37.14 & 29.50 \\
& OmniQ         & 2    & 25.7 & 37.8 & 53.1 & 26.4 & 50.6 & 38.72 & $4.2\times10^4$ \\
& SpinQuant     & 2    & 25.7 & 47.7 & 50.8 & 26.2 & 49.5 & 39.98 & 30.20 \\
& EfficientQAT  & 2.12 & 31.1 & 60.6 & 70.3 & 56.3 & 59.3 & 55.52 & 14.70 \\
& AQLM          & 2.27 & 45.7 & 75.6 & \textbf{78.4} & 73.9 & \textbf{71.6} & 69.04 & \textbf{6.90} \\
& GPTVQ         & 2    & 21.9 & 37.8 & 53.9 & 27.7 & 50.5 & 38.37 & 12.15 \\
& SignRound     & 2    & 33.1 & 64.6 & 69.3 & 56.7 & 60.1 & 56.75 & $4.99\times10^4$ \\
& TesseraQ      & 2    & 22.1 & 57.1 & 53.3 & 54.0 & 58.4 & 48.97 & 100.86 \\
\hline

\end{tabular}
\caption{Main results on LLaMA-3 1B, 3B, and 8B under approximately 2-bit quantization within a practical single-GPU optimization budget. Higher is better ($\uparrow$) except WikiText-2 perplexity ($\downarrow$). All reproduced baselines are run with official or recommended configurations and evaluated using the same benchmark pipeline. We include standard standalone PTQ methods (AWQ, GPTQ, OmniQuant, SpinQuant), QAT baselines (LLM-QAT, EfficientQAT), vector-quantization baselines (AQLM, GPTVQ), and recent ultra-low-bit baselines (SignRound, TesseraQ). Degenerate perplexity values are retained for transparency, but are interpreted as evidence of instability in the direct 2-bit setting rather than as the primary basis for WINDQuant's advantage. Long-budget methods whose recommended optimization cost substantially exceeds the practical-budget setting are discussed separately in Table~\ref{res:runtime}.}
\label{tab:main_results}
\end{table*}

Table~\ref{tab:main_results} reports the main evaluation results under a practical single-GPU optimization budget. This setting reflects the intended use case of WINDQuant: obtaining competitive ultra-low-bit models without relying on extremely long retraining or search procedures. Compared with the previous version, we include additional recent low-bit baselines, including GPTVQ, SignRound, and TesseraQ, to cover vector-quantization, rounding-based, and tensor/block-reconstruction approaches. Methods whose recommended configurations require substantially larger optimization budgets are not omitted from the paper; instead, they are treated as long-budget reference methods and compared separately in Table~\ref{res:runtime}, where runtime and memory cost are shown alongside quality.

\paragraph{Main comparison.}
Table~\ref{tab:main_results} shows that WINDQuant achieves the strongest average downstream performance across all three evaluated scales while keeping WikiText-2 perplexity in a stable regime. On 1B, WINDQuant reaches 49.30 average accuracy, outperforming EfficientQAT (46.58), SignRound (45.09), GPTVQ (41.44), and TesseraQ (39.38). On 3B, WINDQuant obtains 61.98 average accuracy at 1.99 effective bits, slightly above AQLM (61.86 at 2.27 bits) and substantially above SignRound (53.13), TesseraQ (46.13), and EfficientQAT (44.80). On 8B, WINDQuant reaches 70.22 average accuracy at 1.96 bits, outperforming AQLM (69.04 at 2.27 bits), EfficientQAT (55.52), SignRound (56.75), TesseraQ (48.97), and GPTVQ (38.37).

\paragraph{Interpretation.}
The degenerate perplexities of several standalone PTQ baselines should be interpreted as evidence that direct 2-bit PTQ is unstable, not as the sole basis for WINDQuant's advantage. The more informative comparison is against recent low-bit and vector-quantization baselines. AQLM is the closest competitor, achieving slightly better WikiText-2 perplexity on 8B, but WINDQuant obtains higher average accuracy at a lower effective bit-width. SignRound reaches competitive downstream scores on some tasks, but its WikiText-2 perplexity is highly unstable under our reproduced 2-bit setting. Overall, these results suggest that WINDQuant's learned global allocation policy provides a favorable compression--accuracy trade-off by deciding how a limited precision budget should be distributed across the model, rather than only improving local quantization of individual groups.

\begin{table}[t]
  \centering
  \small
  \setlength{\tabcolsep}{3.5pt}
  \begin{tabular}{l  |c c  |c c}
    \hline
    \textbf{Method} & \textbf{Time} & \textbf{GPU Mem.} & \textbf{Avg$\uparrow$} & \textbf{Wiki2 $\downarrow$}  \\
    \hline
    WINDQuant  & 49 & 17.1 & 70.2 & \textbf{7.9}  \\
    LLM-QAT  & 0.5 & 126.9 & 42.0 & 7137   \\
    EfficientQAT & 14 & 8.1 & 44.8 & 14.7   \\
    ParetoQ$^\dagger$ & 1777 & 82.4 & \textbf{70.9} & 8.0  \\
    \hline
  \end{tabular}
  \caption{Efficiency comparison on LLaMA-3.1-8B under approximately 2-bit quantization. WINDQuant and reproduced baselines are measured in our environment. $^\dagger$ ParetoQ is included as a long-budget reference rather than a practical-budget baseline: its quality numbers are taken from the original paper \citep{LiuZechun2025PISL}, while its optimization cost is estimated from running the recommended configuration on our single-H200 setup. Runtime is reported in hours and GPU memory in GB.}
  \label{res:runtime}
\end{table}

\paragraph{Efficiency comparison.}
Table~\ref{res:runtime} compares quality, optimization time, and peak memory consumption on LLaMA-3.1-8B. This comparison is important because ultra-low-bit quantization is not only an accuracy problem: methods that require extremely long optimization may be impractical for iterative compression or deployment-oriented workflows.

WINDQuant requires 49 hours and 17.1 GB of peak GPU memory while achieving 70.2 average accuracy and a WikiText-2 perplexity of 7.9, giving the best quality--efficiency trade-off among the reproduced practical-budget baselines. LLM-QAT is faster in wall-clock time, but requires 126.9 GB of memory and produces a degenerate perplexity of 7137. EfficientQAT has lower memory usage and shorter runtime, but its average accuracy remains substantially lower and its perplexity is nearly twice that of WINDQuant.

ParetoQ is included as a long-budget reference rather than in the practical-budget main table. Although it reports comparable quality, with 70.9 average accuracy and 8.0 WikiText-2 perplexity, it requires 1777 hours and 82.4 GB of memory under its recommended configuration. This supports the central claim of WINDQuant: learned fine-grained allocation can approach the quality of much heavier ultra-low-bit optimization methods while remaining substantially cheaper to optimize.

The computational profile of WINDQuant is dominated by policy rollout, lightweight local quantizer fitting, and calibration-based proxy evaluation rather than full model retraining, avoiding global backpropagation through the full model and reducing optimization cost relative to retraining-heavy alternatives.

\paragraph{Quantization on LLaMA-3-70B.}

\begin{table}[t]
\small
\centering
\setlength{\tabcolsep}{5pt}
\begin{tabular}{l|c|cc}
\hline
\textbf{Method} & \#Bits  
& \textbf{Avg} $\uparrow$ 
& \textbf{Wiki2} $\downarrow$ \\
\hline
FP & 16  
& \textbf{77.28} & \textbf{2.53} \\
\hline
WINDQuant & 2.02  
& \textbf{71.48} & 5.10 \\
AQLM & 2.27 
& 70.04 & \textbf{4.57} \\
TesseraQ & 2.00 
& 66.78 & 7.47 \\
EfficientQAT & 2.12 
& 38.90 & $1.2\times10^{7}$ \\
\hline
\end{tabular}
\caption{Results on LLaMA-3-70B under approximately 2-bit quantization. We attempted WINDQuant together with several baselines that were feasible at smaller scales. In our single-H200 setup, WINDQuant completed with stable quality, while other attempted baselines either failed to maintain a usable perplexity regime or did not complete within the available memory and runtime budget.}
\label{tab:70b_results}
\end{table}

Quantization at the 70B scale presents a significantly more challenging regime, where conventional PTQ methods degrade severely under ultra-low-bit constraints and several stronger baselines become difficult to run within a single-GPU budget. We therefore restrict the 70B comparison to methods that completed with reportable outputs in our environment or had directly comparable available results. As shown in Table~\ref{tab:70b_results}, WINDQuant achieves 71.48 average accuracy at 2.02 bits, preserving more than 92\% of the full-precision score while keeping WikiText-2 perplexity at 5.10. AQLM obtains slightly lower average accuracy (70.04) at a higher effective bit-width of 2.27 bits, while EfficientQAT fails to maintain a usable language-modeling regime under our reproduced setting, with WikiText-2 perplexity increasing to $1.2\times10^7$. These results suggest that the learned allocation policy remains effective at larger model scales, where fine-grained precision assignment can exploit the redundancy of high-capacity models while avoiding the cost of full retraining.


\section{Conclusion}

In this work, we introduced WINDQuant, a reinforcement learning framework for ultra-low-bit mixed-precision quantization of large language models. By formulating precision allocation as a sequential decision-making problem, WINDQuant assigns bit-widths to fine-grained column-chunk units under a global storage budget. The framework combines PPO-based allocation, activation-aware calibration, lightweight local quantizer fitting, and explicit effective-bit accounting to produce learned mixed-precision plans. Across LLaMA models from 1B to 70B parameters, WINDQuant achieves a competitive compression--quality trade-off while avoiding full model retraining and substantially reducing optimization cost compared with heavier training-based approaches. These results suggest that reinforcement learning is a promising direction for adaptive, budget-aware model compression.

\section*{Limitations}

This work focuses on weight-only ultra-low-bit quantization. We do not quantize activations or KV cache states, so the reported bit-widths reflect weight-storage cost rather than full serving memory or latency.

WINDQuant uses a lightweight calibration proxy to train the RL policy. This makes policy optimization practical at billion-parameter scale, but the proxy may not fully capture downstream task accuracy, long-context behavior, or domain shift across calibration corpora. Our proxy analysis suggests that it tracks final WikiText-2 perplexity along the training trajectory, but broader validation across datasets and models remains necessary.

The current pipeline also includes practical stabilizers such as activation-aware salient weight protection, local per-unit scale fitting, fallback rules, and budget-aware action masking. These components are important for robustness in the ultra-low-bit regime, but they make WINDQuant a complete allocation-and-quantization pipeline rather than a pure RL policy alone. More extensive ablations over state features, reward components, and protection mechanisms would further clarify the contribution of each component.

Finally, experiments are primarily conducted on LLaMA-family models in a single-GPU setting. Although we evaluate scales from 1B to 70B parameters, additional architectures, calibration corpora, and deployment environments are needed to establish broader generality.

\bibliography{custom}
\clearpage
\newpage
\appendix

\begin{table*}[t]
\centering
\small
\setlength{\tabcolsep}{4pt}
\begin{tabular}{l|c|ccccc|cc}
\hline
\textbf{Method} & \#Bits 
& \textbf{ARC-e} $\uparrow$
& \textbf{ARC-c} $\uparrow$
& \textbf{PIQA} $\uparrow$
& \textbf{HS} $\uparrow$
& \textbf{Wino} $\uparrow$
& \textbf{Avg} $\uparrow$
& \textbf{Wiki2} $\downarrow$ \\
\hline
FP & 16 
& 86.6 & 58.6 & 83.8 & 85.6 & 81.1 & 79.14 & 2.53 \\
\hline
WINDQuant & 2.02 
& \textbf{80.3} & \textbf{51.2} & \textbf{81.7} & \textbf{78.3} & 75.8 & \textbf{73.46} & 5.10 \\
\hline
AQLM & 2.27 
& 79.0 & 50.1 & 78.1 & 63.5 & \textbf{79.5} & 70.04 & \textbf{4.57} \\
EfficientQAT & 2.12 
& 26.5 & 38.0 & 53.7 & 26.4 & 49.0 & 38.72 & $1.2\times10^{7}$ \\
TesseraQ & 2.00 
& 78.7 & 47.4 & 78.2 & 57.9 & 71.7 & 66.78 & 7.47 \\
\hline
\end{tabular}
\caption{Full per-task results on LLaMA-3-70B under $\sim$2-bit quantization. Higher is better ($\uparrow$) except WikiText-2 perplexity ($\downarrow$). 
}
\label{tab:70b_full}
\end{table*}

\section{Experiment details}
\subsection{Complete results for LLaMa-3-70B}
\label{sec:ap2}

Table~\ref{tab:70b_full} provides the complete per-task results for LLaMA-3-70B under approximately 2-bit quantization.

WINDQuant achieves strong and consistent performance across all tasks at the 70B scale. In particular, it remains closest to full precision on PIQA (81.7 vs.\ 83.8) and HellaSwag (78.3 vs.\ 85.6), indicating that commonsense reasoning is largely preserved under extreme quantization. This suggests that a significant portion of high-level semantic reasoning can be maintained even when the majority of weights operate at ultra-low precision.

On more challenging benchmarks such as ARC-c, WINDQuant reaches 51.2 compared to 58.6 in full precision. The larger gap on ARC-c indicates that tasks requiring multi-step or structured reasoning remain more sensitive to aggressive compression, even at the 70B scale.

Overall, WINDQuant achieves an average accuracy of 73.46, maintaining most of the full-precision performance (79.14) at only 2.02 bits. This result highlights that the proposed policy scales effectively to high-capacity models, where redundancy in parameter space can be better exploited through fine-grained precision allocation.

In terms of language modeling quality, WINDQuant attains a WikiText-2 perplexity of 5.10, which remains close to the full-precision baseline (2.53). In contrast, EfficientQAT exhibits extremely high perplexity ($1.2\times10^{7}$), indicating a failure to preserve model quality under ultra-low-bit constraints. This further reinforces that maintaining a stable perplexity regime is critical for scaling quantization methods to large models.



\subsection{Baseline Configuration Details}
\label{sec:baseline_configs}

To ensure a fair evaluation, all baselines were configured following the recommendations in their respective publications and official repositories whenever possible. Table~\ref{tab:baseline_configs} summarizes the exact hyperparameters used in our experiments. Unless otherwise stated, all methods quantize linear weight matrices only while excluding the embedding and output head layers for stability. For weight-only PTQ and vector-quantization baselines, we evaluate approximately 2-bit weight quantization under each method's intended configuration.

\begin{table*}[t]
\centering
\small
\setlength{\tabcolsep}{3pt}
\begin{tabular}{l|c|c|c|c|p{6.0cm}}
\hline
\textbf{Method} & \textbf{Group} & \textbf{Calib.\ Dataset} & \textbf{Calib.\ $n$} & \textbf{Seq.\ Len} & \textbf{Key Hyperparameters} \\
\hline
AWQ & 128 & Open-Platypus & 512 & 2048 
& Asym.\ quantization at 2--3 bit; sym.\ at $\ge$4 bit \\

GPTQ & 128 & Open-Platypus & 512 & 2048 
& Asym.\ quantization at 2--3 bit; Hessian-weighted rounding \\

OmniQuant & 128 & WikiText-2 & 128 & 2048 
& LWC only; 10 epochs (5 for 8B); LR 0.01 \\

SpinQuant & 128 & C4 & 128 & 2048 
& R3 learned rotations; W2A16KV16 \\

SignRound & 128 & NeelNanda/pile-10k & 128 & 2048 
& 2-bit weight-only; 200 iterations; asym=True; batch size 8; AMP=True \\

TesseraQ & 128 & WikiText-2 & 128 & 2048 
& 2-bit weight-only; w2g128 config; reduce\_memory=True; no weight\_clip; no load\_transform \\
\hline
LLM-QAT & --- & Synthesized & --- & 2048 
& W2A8KV4; 1 epoch; LR 2e-5; cosine schedule; KD \\

EfficientQAT & 64 & Open-Platypus & 1024 & 1024/2048 
& Block-AP 1 ep + E2E-QP 5 ep on Deita-10K \\
\hline
AQLM & 8 & WikiText-2 & 1024 & 4096 
& 1$\times$16 codebook; 10 epochs; early-stop 3 \\

GPTVQ & 256 & WikiText-2 & 128 & 2048 
& 2-bit weight-only; vq-dim=1; codebook-bits=8; kmeans-iters=100; Hessian-weighted lookups; M-step; Mahalanobis initialization \\
\hline
\end{tabular}
\caption{Exact configurations used for the baseline methods in our experiments. ``Group'' denotes the quantization group size when applicable. ``Calib.\ $n$'' denotes the number of calibration samples. The upper block contains standard and recent PTQ baselines, the middle block contains QAT baselines, and the lower block contains vector-quantization baselines. RTN and QuIP are omitted from the main configuration table because they are reported only in the extended appendix comparison.}
\label{tab:baseline_configs}
\end{table*}

\paragraph{PTQ baselines.}
AWQ and GPTQ were run through the \texttt{llmcompressor} one-shot pipeline with asymmetric 2-bit quantization and group size 128, using 512 Open-Platypus calibration samples at sequence length 2048, following the recommended settings of \citet{lin2024awq} and \citet{frantar2022gptq}. OmniQuant uses learnable weight clipping (LWC) only, optimized on 128 WikiText-2 samples at sequence length 2048 for 10 epochs, with the 8B run reduced to 5 epochs due to runtime constraints, following \citet{shao2023omniquant}. SpinQuant follows its two-stage procedure: it first learns R3 orthogonal rotations on 128 C4 samples and then applies W2A16 post-training quantization with group size 128, consistent with \citet{liu2024spinquant}.

We additionally include recent ultra-low-bit baselines that are closer to the target regime of this paper. SignRound is evaluated as a 2-bit weight-only baseline with group size 128, 128 calibration samples from \texttt{NeelNanda/pile-10k}, sequence length 2048, 200 optimization iterations, asymmetric quantization, batch size 8, and AMP enabled. TesseraQ is evaluated using its 2-bit weight-only w2g128 configuration with 128 WikiText-2 calibration samples at sequence length 2048, \texttt{reduce\_memory=True}, no weight clipping, and no loaded transform. These baselines are included to strengthen the comparison beyond conventional PTQ recipes and to cover recent rounding-based and tensor/block-reconstruction approaches in the ultra-low-bit setting.

\paragraph{QAT baselines.}
EfficientQAT follows its two-stage training protocol: Block-AP with 1024 Open-Platypus samples at sequence length 1024, followed by E2E-QP on 2048 Deita-10K samples at sequence length 2048 with learning rate 2e-5 and gradient accumulation 16 (32 for 8B), matching \citet{chen2025efficientqat}. LLM-QAT uses W2A8KV4 quantization with 1 epoch of training on synthesized data, learning rate 2e-5, cosine scheduling, and knowledge distillation, consistent with \citet{liu2024llm}. We retain LLM-QAT in the main comparison because it represents an established QAT-style low-bit LLM compression baseline, while EfficientQAT provides a stronger recent QAT reference.

\paragraph{Vector quantization baselines.}
AQLM uses a 1$\times$16 codebook with in-group size 8, trained for 10 epochs with early stopping patience 3 on 1024 WikiText-2 samples at sequence length 4096, following \citet{egiazarian2024extreme}. GPTVQ is evaluated as a 2-bit weight-only vector-quantization baseline with group size 256, 128 WikiText-2 calibration samples at sequence length 2048, vq-dim=1, codebook-bits=8, 100 k-means iterations, Hessian-weighted lookups, M-step refinement, and Mahalanobis initialization. Including both AQLM and GPTVQ allows us to compare WINDQuant against two vector-quantization approaches with different optimization and codebook designs.

\paragraph{Regarding unusually high perplexity of PTQ baselines.}
We emphasize that the poor performance of standard PTQ methods (AWQ, GPTQ, RTN, SpinQuant, QuIP) at 2-bit is consistent with findings in prior work \cite{ egiazarian2024extreme}: these methods were primarily designed and validated at 3--4 bit precision, and their quality collapses rapidly below 3 bits. Our SpinQuant results (e.g., Wiki2 PPL of 56.34 on 1B, 57.4 on 3B) are broadly consistent with the ranges reported in the SpinQuant paper for 2-bit settings. RTN, AWQ, and GPTQ produce degenerate outputs ($10^4$--$10^6$ perplexity) because their quantization noise overwhelms the signal at 2-bit group-128 granularity. No additional tuning was performed beyond the official recommended configurations, as our goal was to evaluate each method under its standard settings.

\subsection{Learned policy behavior}
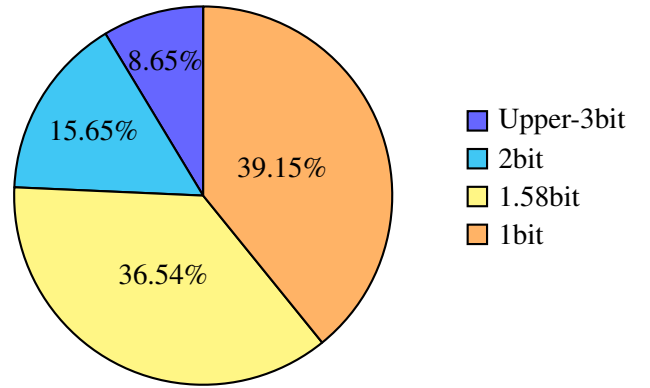
\begin{figure}[t]
\centering
\begin{tikzpicture}
\pie[
    text=legend,
    radius=2.5,
    after number=\%,
    rotate=90,
]{
    8.65/Upper-3bit,
    15.65/2bit,
    36.54/1.58bit,
    39.15/1bit
}
\end{tikzpicture}
\caption{Action distribution of the RL quantization policy when quantizing LLaMA-3.1-8B. Higher-bit actions ($\ge$3 bits and skip) are grouped into ``Upper-3bit'', showing that the policy strongly favors ultra-low-bit configurations (1--2 bits).}
\label{fig:action_distribution_pie_grouped}
\end{figure}

As shown in Figure~\ref{fig:action_distribution_pie_grouped}, the RL agent strongly favors ultra-low-bit configurations, with approximately 75\% of weights assigned to 1-bit and 1.58-bit, and only a small fraction (8.65\%) allocated to higher precision ($\ge$3 bits). This distribution indicates that WINDQuant does not simply preserve many units at high precision. Instead, it learns an aggressive allocation policy that assigns most units to binary or ternary-like regimes while reserving higher precision for a small subset of more sensitive units. This behavior aligns with the core design goal of WINDQuant: exploiting fine-grained redundancy through learned allocation rather than relying on uniform or layer-wise precision choices.

\section{Design Choices and Empirical Justification}
\subsection{Quality vs.\ Bit-width Trade-offs}
\label{sec:ap4}

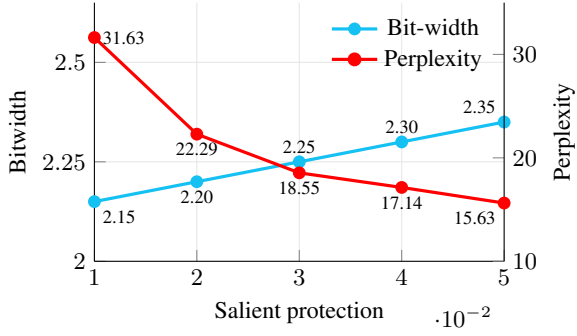
\begin{figure}[t]
\centering
\small
\begin{tikzpicture}
\begin{axis}[
    width=7cm,
    height=5cm,
    xlabel={Salient protection},
    ylabel={Bitwidth},
    xmin=0.01, xmax=0.05,
    ymin=2.0, ymax=2.65,
    xtick={0,0.01,0.02,0.03,0.04,0.05},
    ytick={2.0,2.25,2.5},
    legend style={at={(0.98,0.98)}, anchor=north east, draw=none, fill=none},
    axis y line*=left,
    axis x line*=bottom,
    grid=both,
    grid style={gray!20},
    tick label style={font=\small},
    label style={font=\small},
]

\addplot[
    color=cyan!70,
    line width=1.2pt,
    mark=*,
    mark size=1.8pt,
] coordinates {
    (0.01,2.15)
    (0.02,2.20)
    (0.03,2.25)
    (0.04,2.30)
    (0.05,2.35)
};

\addlegendentry{Bit-width}
\addlegendimage{color=red!70!red, line width=1.2pt, mark=*}
\addlegendentry{Perplexity}

\node[font=\scriptsize] at (axis cs:0.01,2.15) [below right] {2.15};
\node[font=\scriptsize] at (axis cs:0.02,2.20) [below ] {2.20};
\node[font=\scriptsize] at (axis cs:0.03,2.25) [above ] {2.25};
\node[font=\scriptsize] at (axis cs:0.04,2.30) [above ] {2.30};
\node[font=\scriptsize] at (axis cs:0.05,2.35) [above left] {2.35};

\end{axis}

\begin{axis}[
    width=7cm,
    height=5cm,
    xmin=0.01, xmax=0.05,
    ymin=10, ymax=35,
    ylabel={Perplexity},
    ytick={10,20,30},
    axis y line*=right,
    axis x line=none,
    tick label style={font=\small},
    label style={font=\small},
]

\addplot[
    color=red!70!red,
    line width=1.2pt,
    mark=*,
    mark size=1.8pt,
] coordinates {
    (0.01,31.63)
    (0.02,22.29)
    (0.03,18.55)
    (0.04,17.14)
    (0.05,15.63)
};

\node[font=\scriptsize] at (axis cs:0.01,31.63) [right] {31.63};
\node[font=\scriptsize] at (axis cs:0.02,22.29) [below] {22.29};
\node[font=\scriptsize] at (axis cs:0.03,18.55) [below] {18.55};
\node[font=\scriptsize] at (axis cs:0.04,17.14) [below] {17.14};
\node[font=\scriptsize] at (axis cs:0.05,15.63) [below left] {15.63};

\end{axis}
\end{tikzpicture}
\caption{Effect of salient protection on average bit-width and WikiText-2 perplexity for 5-episode 2-bit quantization of LLaMA-3.2-1B. Increasing salient protection raises the effective bit-width while generally improving perplexity.}
\label{fig:salient_protection_dual_axis}
\end{figure}

As shown in Figure~\ref{fig:salient_protection_dual_axis}, increasing the salient protection ratio leads to a clear trade-off between effective bit-width and language modeling quality. When the protection ratio is very low (e.g., 0.01), the model operates under stricter compression (around 2.15 bits) but suffers from significantly degraded perplexity (31.63). As the protection ratio increases, perplexity improves rapidly, dropping to 22.29 at 0.02 and further to 18.55 at 0.03. Beyond this point, the marginal improvement in perplexity diminishes while the effective bit-width continues to rise, reaching 2.35 at 0.05 in the figure. Based on this observation, we choose a protection ratio of 0.03 as a balanced operating point: it significantly improves perplexity compared to more aggressive compression, while keeping the bit-width close to the target ultra-low regime. This analysis is intended to select a stable protection rate rather than to isolate the contribution of the RL policy. Salient protection alone only determines a small set of INT8-preserved weights; the precision of the remaining majority of parameters is still decided by the learned allocation policy. We therefore combine this protection-rate study with the heuristic allocator comparison in Section~\ref{sec:heuristic_ablation}, where the same protection mechanism and quantization operators are kept fixed while the PPO policy is replaced by a deterministic sensitivity rule.

\subsection{Quality vs. Run-time Trade-offs}


\begin{figure}[t]
\centering
\small
\begin{tikzpicture}
\begin{axis}[
width=7cm,
height=5cm,
xlabel={Episodes},
ylabel={Perplexity},
xmin=50, xmax=350,
ymin=16.4, ymax=20.5,
xtick={50,100,150,200,250,300,350},
ytick={17,18,19,20},
legend style={at={(0.98,0.98)}, anchor=north east, draw=none, fill=none},
axis y line*=left,
axis x line*=bottom,
grid=both,
grid style={gray!20},
tick label style={font=\small},
label style={font=\small},
]

\addplot[
color=red!70,
line width=1.2pt,
mark=*,
mark size=1.8pt,
] coordinates {
(50,20.03)
(100,18.74)
(150,17.27)
(200,17.22)
(250,16.58)
(300,17.22)
(350,17.22)
};

\addlegendentry{Perplexity}

\node[font=\scriptsize] at (axis cs:50,20.03) [above right] {20.03};
\node[font=\scriptsize] at (axis cs:100,18.74) [below left] {18.74};
\node[font=\scriptsize] at (axis cs:150,17.27) [below left] {17.27};
\node[font=\scriptsize] at (axis cs:200,17.22) [above left] {17.22};
\node[font=\scriptsize] at (axis cs:250,16.58) [above] {16.58};
\node[font=\scriptsize] at (axis cs:300,17.22) [above left] {17.22};
\node[font=\scriptsize] at (axis cs:350,17.22) [above left] {17.22};

\end{axis}
\end{tikzpicture}
\caption{Evolution of WikiText-2 perplexity over RL training episodes. Perplexity decreases rapidly during the early stage of training and shows limited additional benefit after 200 episodes, indicating convergence of the learned allocation policy.}
\label{fig:episode_saturation}
\end{figure}
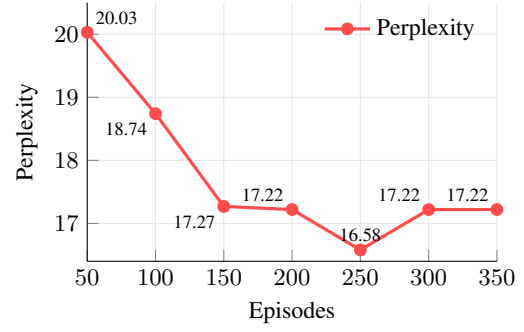

\begin{table}[t]
\centering
\small
\setlength{\tabcolsep}{3.5pt}
\begin{tabular}{l|ccccccc}
\hline
\textbf{Eps} & 50 & 100 & 150 & 200 & 250 & 300 & 350 \\
\hline
\textbf{Wik2} & 20.03 & 18.74 & 17.27 & 17.22 & 16.58 & 17.22 & 17.22 \\
\textbf{Prxy} & 10.19 & 9.86 & 9.29 & 9.29 & 9.06 & 9.28 & 9.27 \\
\hline
\end{tabular}
\caption{Final WikiText-2 perplexity and training-time proxy perplexity across RL training episodes.}
\label{tab:episode_saturation}
\end{table}

Table ~\ref{tab:episode_saturation} and Figure~\ref{fig:episode_saturation} show that the optimization process converges rapidly within the first 200 episodes. WikiText-2 perplexity decreases from 20.03 at 50 episodes to 18.74 at 100 episodes, and further to 17.27 at 150 episodes. At 200 episodes, perplexity reaches 17.22, after which no consistent improvement is observed. Although a lower value of 16.58 appears at 250 episodes, this gain is not sustained, as perplexity returns to 17.22 at 300 and 350 episodes. This indicates that the policy has effectively converged by 200 episodes, and later improvements are not stable. Across checkpoints from 50 to 350 episodes, the training-time proxy is strongly correlated with final WikiText-2 perplexity (Pearson $r=0.997$, Spearman $\rho=0.935$). Although this analysis is limited to one training trajectory, it suggests that the proxy provides a useful signal for guiding policy optimization.

From an efficiency standpoint, extending training beyond 200 episodes is not justified. Under the default setting, every additional 50 episodes increases runtime by approximately 7 hours. Increasing the budget from 200 to 350 episodes therefore adds 21 hours of computation without yielding a reliable improvement in perplexity. Even the extension from 200 to 250 episodes incurs an additional 7 hours while failing to produce a consistent gain. These results support selecting 200 episodes as the default training budget, as it captures the majority of the performance improvement while avoiding unnecessary computational overhead.

\subsection{Comparison with Heuristic Sensitivity Allocator}
\label{sec:heuristic_ablation}

A natural question is whether the RL-learned policy provides value beyond a well-designed deterministic heuristic. To investigate this, we implement a heuristic sensitivity allocator that uses the identical infrastructure as WINDQuant---the same activation-aware saliency protection (3\% salient rate), the same per-unit quantizers (elastic binarization, SEQ, LSQ), and the same effective-bit accounting---but replaces the PPO policy with a deterministic rule.

\paragraph{Heuristic rule.} For each unit, we compute a saliency score as the mean product of activation scale and weight magnitude. Units are ranked by saliency (descending), and bits are assigned greedily: starting from a 2-bit default for all units, the least sensitive units are pushed to 1.58-bit or 1-bit to free budget, and the most sensitive units are promoted to 3-bit or 4-bit if budget permits. The target average bit-width is 2.0.

\paragraph{Experimental setup.} Both WINDQuant (50 episodes) and the heuristic allocator are run on LLaMA-3.2-1B with 5 random seeds $\{42, 123, 456, 789, 2026\}$ to estimate variance. We report mean $\pm$ standard deviation of proxy perplexity and average bit-width.

\paragraph{Discussion.}
The results in Table~\ref{tab:heuristic_vs_rl} show that WINDQuant consistently outperforms the deterministic heuristic under an identical quantization pipeline, indicating that the improvement is not attributable to the underlying quantizers, saliency definition, or protection mechanism, but rather to the learned allocation policy. While the heuristic captures a reasonable first-order approximation of sensitivity by ranking units based on saliency, it remains fundamentally myopic: decisions are made greedily and independently, without accounting for global budget interactions or downstream effects of earlier allocations.

In contrast, the RL policy explicitly optimizes a sequential objective under a global bit constraint, enabling it to discover non-trivial allocation patterns that better balance compression and quality. In particular, the policy can implicitly trade off local sensitivity against global budget pressure, selectively allocating higher precision only when it yields sufficient marginal benefit, while compensating with more aggressive compression elsewhere. The heuristic allocator can approach WINDQuant in proxy perplexity only by using a noticeably larger effective bit-width. WINDQuant achieves a comparable perplexity at roughly 0.12 fewer bits, indicating that the RL policy learns a more efficient allocation under the global budget. This behavior cannot be replicated by a static ranking-based heuristic without additional tuning or hand-crafted rules.

Moreover, the near-zero variance observed across seeds suggests that the learned policy is stable and not sensitive to stochastic training noise in this setting. Overall, these results support the claim that the advantage of WINDQuant arises from its ability to perform coordinated, budget-aware allocation, rather than from auxiliary components shared with the heuristic baseline.

\begin{table}[t]
\centering
\small
\setlength{\tabcolsep}{5pt}
\begin{tabular}{lcc}
\hline
\textbf{Method} & \textbf{Bitwidth} & \textbf{Perplexity}  \\
\hline
Heuristic   & $2.182$ & $8.875$  \\
WINDQuant    & $2.066 \pm 0.001$& $8.756 \pm 0.055$\\
\hline
\end{tabular}
\caption{Comparison of WINDQuant (RL) vs.\ deterministic heuristic allocator on LLaMA-3.2-1B at $\sim$2-bit, averaged over 5 seeds. Both methods use identical saliency protection, quantizers, and bit accounting. WINDQuant involve 50 episodes in the quantization process.}

\label{tab:heuristic_vs_rl}
\end{table}

\subsection{Sensitivity to Chunk Size}
\label{sec:chunk_ablation}

The column-chunk size $G$ controls the granularity of the quantization units. Smaller $G$ creates more units per layer, enabling finer-grained precision allocation at the cost of more RL decisions per episode. Larger $G$ reduces overhead but limits the agent's ability to adapt to within-layer sensitivity variations.

We evaluate three chunk sizes---$G \in \{128, 256, 512\}$---on LLaMA-3.2-1B with 100 episodes and identical hyperparameter (target 2.0-bit, salient rate 0.03, curriculum $\{3.0, 2.5, 2.0, 2.0\}$).

\begin{table}[t]
\centering
\small
\setlength{\tabcolsep}{5pt}
\begin{tabular}{c|ccc|c}
\hline
$G$ & \textbf{Units/ep} & \textbf{PPL} $\downarrow$ & \textbf{BW} & \textbf{Time (h)} \\
\hline
128 & 88 & 8.78 & 2.03 & 20.4 \\
256 & 74 & 8.87 & 2.02 & 14.2 \\
512 & 52 & 10.03 & 2.03 & 7.2 \\
\hline
\end{tabular}
\caption{Effect of chunk size $G$ on proxy perplexity, average bit-width, number of RL decisions per episode, and total runtime for LLaMA-3.2-1B at $\sim$2-bit quantization over 100 episodes.}
\label{tab:chunk_ablation}
\end{table}

As shown in Table~\ref{tab:chunk_ablation}, the chunk size $G$ introduces a clear trade-off between quantization granularity and computational efficiency. A smaller chunk size ($G=128$) provides finer-grained control, achieving the best perplexity (8.78), but incurs the highest runtime (20.4 hours). In contrast, a larger chunk size ($G=512$) reduces the number of RL decisions and total runtime (7.2 hours), but results in a clear degradation in perplexity (10.03), indicating limited capacity to capture within-layer sensitivity variations. 

The intermediate setting $G=256$ provides a balanced trade-off. It achieves a perplexity of 8.87, which is very close to the result at $G=128$ (a difference of 0.09 PPL), while reducing total runtime to 14.2 hours, corresponding to a 30.2\% reduction. This demonstrates that $G=256$ preserves most of the performance benefits of finer granularity while offering a substantial improvement in efficiency, making it a suitable default choice for subsequent experiments.

\section{Implementation Details}

\label{appdx:engine} 
\subsection{Pipeline}
Figure \ref{fig:pipeline} illustrates the overall workflow. Given a pretrained model $\mathcal{M}$, we first decompose each weight matrix into column-level quantization units. At each time step $t$, the environment constructs a state vector $s_t$ summarizing statistical properties of the current unit, structural layer information, and global compression indicators. The RL agent selects an action $a_t$, which is executed by the quantization engine using elastic binarization at 1-bit, SEQ-style operators at 1.58/2-bit, and LSQ-style sparse quantization at 3/4/8-bit, with fallback to a higher precision if reconstruction error is excessive. After each action, a small step reward is recorded. Once all units are processed, a fast perplexity pass on calibration text produces the episode-end reward used by PPO.

\subsection{Per-unit Scale Optimization}
At each step, the selected quantization operator fits its scale parameter by running a small local optimization (typically 10 Adam steps) on the reconstruction loss for the current column chunk. SEQ variants use soft assignments and per-group scales, whereas LSQ-style operators learn step sizes with straight-through estimation for discrete rounding. If the reconstruction error remains too large, the implementation can fall back to the next higher precision. This optimization is entirely local to the unit: it requires no model-wide backward pass and does not interact with the PPO gradient. Group size is set to the chunk size $G{=}256$, matching the unit granularity.


\subsection{Salient Weight Protection}
We adopt activation-aware weight quantization \cite{lin2024awq} to identify and protect salient weights. The saliency of each weight is computed as the product of its magnitude and the corresponding per-channel activation scale, estimated during a calibration forward pass. This measure approximates the contribution of a weight to output variance and prioritizes parameters associated with frequently activated channels.

Given a protection rate $r$ (configured per run; the saved experiments in this repository mainly use fixed salient rates in the 2\%--3\% range), the top-$r$ fraction of weights by saliency are preserved at INT8 precision, while remaining weights use the agent-selected bit-width. The implementation can also suppress protection for tensors that do not satisfy heavy-tail criteria when a fixed salient rate is not supplied.

\paragraph{Effective bit-width accounting.}
The presence of FP16 and INT8-protected parameters means the effective per-unit bit cost differs from the nominal action bit-width. For each unit $i$ with agent-selected action $b_i$, let $n_i^{(p)}$ denote the number of protected parameters and $n_i^{(q)} = n_i - n_i^{(p)}$ the remainder:

\begin{equation}
\label{eq:effective-bits-appdx}
B_i =
\begin{cases}
16 \cdot n_i, & \text{if action = 16}, \\
b_i \cdot n_i^{(q)} + 8 \cdot n_i^{(p)}, & \text{otherwise}.
\end{cases}
\end{equation}

The average bit-width reported in all experiments uses $\bar{b} = \sum_i B_i / \sum_i n_i$, which faithfully reflects the true storage cost including protection overhead.

\end{document}